\documentclass[sigconf,nonacm]{acmart}

\usepackage[T1]{fontenc}
\usepackage[utf8]{inputenc}
\usepackage{amsmath}
\usepackage{algorithm}
\usepackage{algpseudocode}
\usepackage{booktabs}
\usepackage{multirow}
\usepackage{tabularx}
\usepackage{array}
\usepackage{graphicx}
\usepackage{url}

\renewcommand\footnotetextcopyrightpermission[1]{}
\newcolumntype{Y}{>{\centering\arraybackslash}X}
\newcolumntype{L}{>{\raggedright\arraybackslash}X}

\title{Reinforcement Federated Learning Method Based on Adaptive OPTICS Clustering}

\author{Tianyu Zhao}
\affiliation{%
  \institution{Beijing Key Laboratory of Intelligent Communication Software and Multimedia, School of Computer Science (National Pilot Software Engineering School), Beijing University of Posts and Telecommunications}
  \city{Beijing}
  \country{China}}

\author{Junping Du}
\authornote{Corresponding author.}
\email{junpingdu@126.com}
\affiliation{%
  \institution{Beijing Key Laboratory of Intelligent Communication Software and Multimedia, School of Computer Science (National Pilot Software Engineering School), Beijing University of Posts and Telecommunications}
  \city{Beijing}
  \country{China}}

\author{Yingxia Shao}
\affiliation{%
  \institution{Beijing Key Laboratory of Intelligent Communication Software and Multimedia, School of Computer Science (National Pilot Software Engineering School), Beijing University of Posts and Telecommunications}
  \city{Beijing}
  \country{China}}

\author{Zeli Guan}
\affiliation{%
  \institution{Beijing Key Laboratory of Intelligent Communication Software and Multimedia, School of Computer Science (National Pilot Software Engineering School), Beijing University of Posts and Telecommunications}
  \city{Beijing}
  \country{China}}

\begin{abstract}
Federated learning is a distributed machine learning technology, which realizes the balance between data privacy protection and data sharing computing. To protect data privacy, federated learning learns shared models by locally executing distributed training on participating devices and aggregating local models into global models. There is a problem in federated learning, that is, the negative impact caused by the non-independent and identical distribution of data across different user terminals. In order to alleviate this problem, this paper proposes a strengthened federation aggregation method based on adaptive OPTICS clustering. Specifically, this method perceives the clustering environment as a Markov decision process, and models the adjustment process of parameter search direction, so as to find the best clustering parameters to achieve the best federated aggregation method. The core contribution of this paper is to propose an adaptive OPTICS clustering algorithm for federated learning. The algorithm combines OPTICS clustering and adaptive learning technology, and can effectively deal with the problem of non-independent and identically distributed data across different user terminals. By perceiving the clustering environment as a Markov decision process, the goal is to find the best parameters of the OPTICS cluster without artificial assistance, so as to obtain the best federated aggregation method and achieve better performance. The reliability and practicability of this method have been verified on the experimental data, and its effectiveness and superiority have been proved.
\end{abstract}


\begin{document}
\maketitle

\section{Introduction}

With the popularity of mobile computing and the Internet, more and more users are using mobile terminals for data communication and computing. With the increasing demand for user data privacy and local processing, federated learning technology has emerged in this context.

Federated learning technology \cite{mcmahan2017communication,li2022federatedquantization,guan2021federatedgnn,li2026fedsin} is a new distributed machine learning technology proposed in 2017, with Google being one of the first companies to propose this technology. Unlike traditional centralized machine learning, federated learning technology can balance the contradiction between data value and data privacy well. In this distributed computing paradigm, all participating nodes can collaboratively train a global model while protecting user privacy and data security. The data of each participating node is stored locally and can be updated on the server's coordination. The trained model can be allocated to various participants or shared among multiple parties \cite{yang2019ai,sattler2019robust,li2024fedcmr,li2022smcr}. The advantage of this distributed machine learning technology is that it can effectively avoid data leakage \cite{li2022vehicle,shao2021memory,li2022distributedpathqueries} and data abuse caused by data privacy issues \cite{li2021heterogeneouslatenttopic,li2017tobitkalman,kou2018hashtag,wei2019boosting}. At the same time, federated learning technology is widely used in many scenarios, such as medical health, financial technology, intelligent manufacturing, graph retrieval, and recommendation \cite{huang2021hgamn,xiao2022lecf}.

Recent studies also show that federated and graph-based representation learning are useful for heterogeneous information networks and cross-modal scenarios. FedSIN introduces federated self-adaptive learning for information network representation, while federated supervised cross-modal retrieval extends privacy-preserving learning to distributed multimodal retrieval tasks \cite{li2026fedsin,li2024fedcmr}. Scientific and technological information retrieval and publication representation learning further indicate that semantic, media, and high-order hypergraph relations can be jointly exploited for robust scientific data mining \cite{li2022smcr,li2026ssahgc}. These works are related to the motivation of this paper because client representations in federated learning are often heterogeneous, multi-view, and dynamically evolving.

Therefore, solving the problem of non-independent and identically distributed data is one of the important directions for federated learning technology to address challenges \cite{yang2015ontology,lin2009averageconsensus,li2018businesscomputing,li2019interpretabledecision,meng2013tracking}. This paper proposes a reinforcement federated aggregation method based on adaptive OPTICS clustering, which can solve the heterogeneity problem \cite{li2023mvsc,yang2016modularity,hu2019hgat,huo2023t2gnn} of data distribution in federated learning tasks, named FedRO. Compared with traditional federated learning algorithms, the algorithm proposed in this paper can not only achieve efficient data recording, learning, and updating, but also share and aggregate data while protecting user privacy. Multi-view scholar clustering provides evidence that clustering dynamic academic entities is useful for modeling evolving interests, and modularity-based community detection shows that deep learning can assist graph community discovery \cite{li2023mvsc,yang2016modularity}. Heterogeneous graph attention networks and T2-GNN further suggest that incomplete features and heterogeneous graph structures can be handled by attention and teacher-student distillation mechanisms \cite{hu2019hgat,huo2023t2gnn}. In summary, solving the problem of data heterogeneity in federated learning is of great practical significance.

Although recommendation-oriented sequence modeling is not the main focus of this paper, filter-enhanced MLP and self-supervised graph co-training show that lightweight sequence models and graph self-supervision can capture dynamic user or session patterns \cite{zhou2022fmlp,xia2021graphcotrain}. Retrieval-oriented pre-training such as RetroMAE is also related to the construction of discriminative feature representations for downstream retrieval and clustering \cite{xiao2022retromae}. These studies are cited as auxiliary evidence for feature learning and dynamic relation modeling.

The main contributions of this paper are as follows:
\begin{enumerate}
    \item A reinforcement learning-based adaptive OPTICS clustering algorithm is proposed. This algorithm considers the problem of non-independent and identically distributed data across devices in federated learning. By clustering data from different devices, it can effectively solve the problem of uneven data distribution. Specifically, we use the advantages of the OPTICS clustering algorithm and use reinforcement learning to adaptively determine the core distance and minimum sample size of clustering, which can more accurately and reasonably handle data distribution.
    \item A reinforcement federated learning method based on the adaptive OPTICS clustering algorithm is proposed. This method clusters clients into different clusters based on features using the adaptive OPTICS clustering algorithm and performs random selection within the clusters, which can make the federated learning more stable and accurate.
    \item Experimental results show that the proposed adaptive OPTICS clustering algorithm and reinforcement federated aggregation method can effectively address the problem of data heterogeneity in federated learning tasks with non-independent and identically distributed data across different devices, improving the performance and accuracy of federated learning. The algorithm has good performance on the MNIST, CIFAR-10, and Fashion-MNIST datasets.
\end{enumerate}

\section{FedRO}

\begin{figure*}[t]
    \centering
    \includegraphics[width=0.72\textwidth]{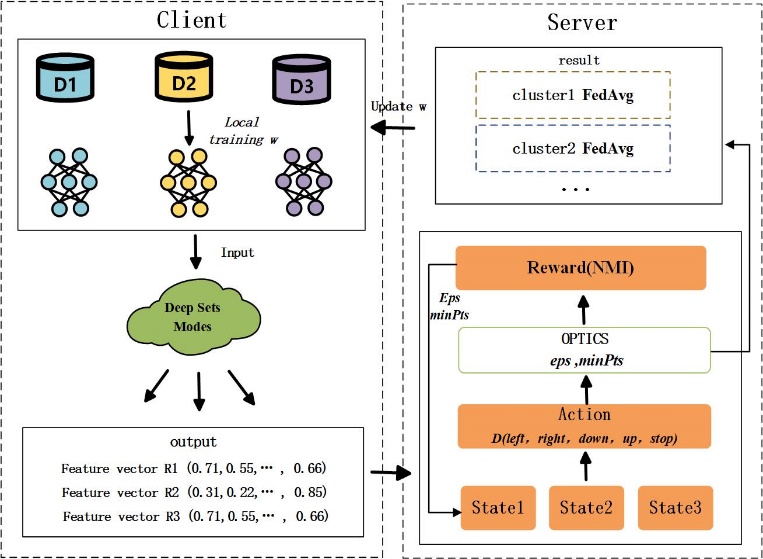}
    \caption{Illustration of FedRO.}
    \Description{The FedRO architecture contains clients that perform local training and feature extraction with Deep Sets, and a server that uses reinforcement learning, rewards, actions, and adaptive OPTICS clustering to form clusters for federated aggregation.}
    \label{fig:fedro}
\end{figure*}

The core idea of this article is to extract features from local data using the Deep Sets model, upload feature vectors to the server node, and use reinforcement learning to define the state space, actions, and rewards. The clustering environment is modeled as a Markov decision process, and the parameter search direction adjustment process is modeled to find the optimal clustering parameters \textit{eps} and \textit{minPts} for the OPTICS clustering algorithm to achieve the best federated aggregation method. Similar clients are assigned to the same cluster, and random selection is performed within the cluster. Each cluster is used to determine a model. The overall architecture of the solution is shown in Figure~\ref{fig:fedro}, which depicts the process of implementing reinforcement federated aggregation using the adaptive OPTICS clustering algorithm with three clients as an example.

\subsection{Reinforced Federated Learning Method}

Specifically, the search process in round $k$ takes the following form. As the state needs to represent the search environment of each step as accurately and completely as possible, we consider constructing the representation of the state from two aspects ($i=1,2,\ldots$). First, the definition of the state for the global clustering status is:
\begin{equation}
    s_{\mathrm{global}}^{(k)(i)} = P^{(k)(i)} \cup D_b^{(k)(i)} \cup \left\{ R_{c_n}^{(k)(i)} \right\} .
\end{equation}

Secondly, for the description of each class, for the local state of class $c_n \in C$ at step $i$, this paper defines a local clustering status definition:
\begin{equation}
    s_{\mathrm{local},n}^{(k)(i)} = \chi_{\mathrm{cent},n}^{(k)(i)} \cup \left\{ D_{\mathrm{cent},n}^{(k)(i)} \mid c_n^{(k)(i)} \right\} .
\end{equation}

Based on the global state and local state, the current state is defined as:
\begin{equation}
    s^{(k)(i)} = \sigma\left(F_G\left(s_{\mathrm{global}}^{(k)(i)}\right) \parallel \sum_{c_n\in C} F_L\left(s_{\mathrm{local},n}^{(k)(i)}\right)\right),
\end{equation}
where $F_G(\cdot)$ and $F_L(\cdot)$ represent global and local feature encoders, respectively, and $\parallel$ denotes concatenation.

\paragraph{Action.} The action represents the parameter search direction for step $i$. This paper defines the action space as $D=\{\textit{left},\textit{right},\textit{down},\textit{up},\textit{stop}\}$, where \textit{left} and \textit{right} represent decreasing and increasing the \textit{eps} parameter, respectively. \textit{Down} and \textit{up} represent decreasing and increasing the \textit{minPts} parameter, respectively, while \textit{stop} represents stopping the search. Specifically, this paper establishes an Actor as a policy network based on the current state:
\begin{equation}
    a^{(k)(i)} = Actor\left(s^{(k)(i)}\right) .
\end{equation}

\paragraph{Reward.} This paper uses a small portion of externally measured samples as the basis for rewards. The reward for step $i$ is:
\begin{equation}
    R\left(s^{(k)(i)}, a^{(k)(i)}\right) = NMI\left(OPTICS(P',y,c)\right) .
\end{equation}

\paragraph{Termination.} For the entire search process, we use the following termination conditions: stop when exceeding the boundary, or stop when exceeding the maximum number of steps.

The specific algorithm is shown in Algorithm~\ref{alg:fedro}.

\begin{algorithm}[t]
\caption{The core model of FedRO}
\label{alg:fedro}
\begin{algorithmic}[1]
\Require Server $S$, clients $C_1,\ldots,C_n$, initial weight $w_i$, learning rate $\eta$.
\Ensure The global parameter models $w_n$ of $n$ clusters.
\State $R_i \gets DeepSets(C_i,w_i)$
\For{$e=1$ to $E_{\max}$}
    \State Initialize $eps^{(e)(0)}$ and $minPts^{(e)(0)}$
    \For{$i=1$ to $I_{\max}$}
        \State Obtain the current state $s^{(e)(i)}$
        \State Choose the action $a^{(e)(i)}$
        \State Get new parameters $eps^{(e)(i)}$ and $minPts^{(e)(i)}$
        \State Make the termination judgment
        \If{is \textsc{Train}}
            \State Get reward $r^{(e)(i)}$
        \EndIf
    \EndFor
    \State Update optimal parameters $eps$ and $minPts$
\EndFor
\State $\Gamma \gets OPTICS(R_1,\ldots,R_n)$
\For{each cluster $(c_1,c_2,\ldots)\in \Gamma$}
    \State $S_t \gets$ a random set of $n$ clients
    \For{each client $c\in S_t$}
        \State $ClientUpdate(k,w_t)$
    \EndFor
    \State $w_{t+1}\gets \sum_{k=1}^{K} \frac{n_k}{n} w_{t+1}^{k}$
\EndFor
\State \textbf{ClientUpdate}$(k,w)$:
\For{each local epoch $I$ from 1 to $E$}
    \For{batch $b\in \beta$}
        \State $w \gets w - \eta \nabla \ell(w;b)$
    \EndFor
\EndFor
\State \Return $w$
\end{algorithmic}
\end{algorithm}

The feature matrix is clustered using adaptive parameters with OPTICS, and the nodes are grouped. The weights of the entire cluster model are then sent to the worker nodes. The FedAvg algorithm is used to locally update the cluster, and the weights are sent back to the server. Finally, the server receives all the weights and aggregates them with weights.

\section{Experiments}

To verify the effectiveness of the FedRO algorithm, this paper evaluates the algorithm by training popular CNN models on three datasets, namely MNIST, CIFAR-10 \cite{krizhevsky2009learning}, and Fashion-MNIST \cite{xiao2017fashionmnist}.

\subsection{Datasets and Models}

The statistical information of the experimental datasets and models is shown in Table~\ref{tab:datasets-models}. To demonstrate the reliability of the adaptive clustering algorithm in this paper, four datasets, glass, wine, yeast, and iris, are selected to test the clustering algorithm in this paper, using two evaluation indicators, Adjusted Rand Index (ARI) and Normalized Mutual Information (NMI). The specific information of the datasets is shown in Table~\ref{tab:data-description}.

\begin{table*}[t]
\centering
\caption{Data set and model statistics.}
\label{tab:datasets-models}
\begin{tabularx}{\textwidth}{L Y Y}
\toprule
Dataset & Number of samples & Model \\
\midrule
MNIST & 50000 & CNN \\
CIFAR-10 & 60000 & CNN \\
Fashion-MNIST & 60000 & CNN \\
\bottomrule
\end{tabularx}
\end{table*}

\begin{table*}[t]
\centering
\caption{Data description.}
\label{tab:data-description}
\begin{tabularx}{\textwidth}{L Y Y Y}
\toprule
Dataset & Samples & Features & Classes \\
\midrule
glass & 214 & 9 & 6 \\
wine & 178 & 13 & 3 \\
yeast & 1484 & 8 & 9 \\
iris & 150 & 4 & 3 \\
\bottomrule
\end{tabularx}
\end{table*}

\subsection{Experiment Settings}

In this paper, experiments were conducted on different datasets using 100 available devices. The datasets were assigned to different training nodes according to different sampling methods, and each node trained its own data locally. After each round of training, these training nodes sent their training results, i.e., their own model parameters, to the parameter server. The parameter server used a certain algorithm to aggregate these model parameters to obtain a new global model parameter. Then, this new global model parameter was distributed to the next round of training nodes, allowing them to continue training their local data using the latest global model parameter.

The following methods were selected as benchmarks for comparative experiments in this study: FedAvg, FedProx \cite{li2020fedprox}, and SHARE \cite{deng2021share}. The FedProx algorithm is an optimization aggregation algorithm proposed in 2020 to solve system and statistical heterogeneity in federated networks, and SHARE is a hierarchical federated learning method proposed in 2021.

In federated learning, reducing the number of communication rounds is important due to the limited computing power and network bandwidth of mobile devices. Therefore, we use the number of communication rounds and the highest accuracy as performance indicators. In the clustering experiment, on the datasets shown in Table~\ref{tab:datasets-models}, the clustering indicators of the four clustering algorithms, K-means, DBSCAN, OPTICS, and the reinforcement learning-based adaptive OPTICS algorithm proposed in this paper, were compared. The ARI and NMI of the four algorithms were compared.

\subsection{Federated Learning Experiment Results and Analysis}

In reference \cite{wang2020optimizing}, the parameter $\sigma$ is used to measure the degree of non-independent and identically distributed data. We also use this method in this article, where a larger $\sigma$ indicates a stronger degree of non-independent and identically distributed data. We use two parameters, 0.8 and 1.0, respectively. When $\sigma=1.0$, it means that the data is the most non-independent and identically distributed, and each label belongs to only one device. When $\sigma=0.8$, it means that 80\% of the data belongs to one label and the remaining 20\% belongs to other labels.

\begin{figure*}[t]
    \centering
    \includegraphics[width=\textwidth]{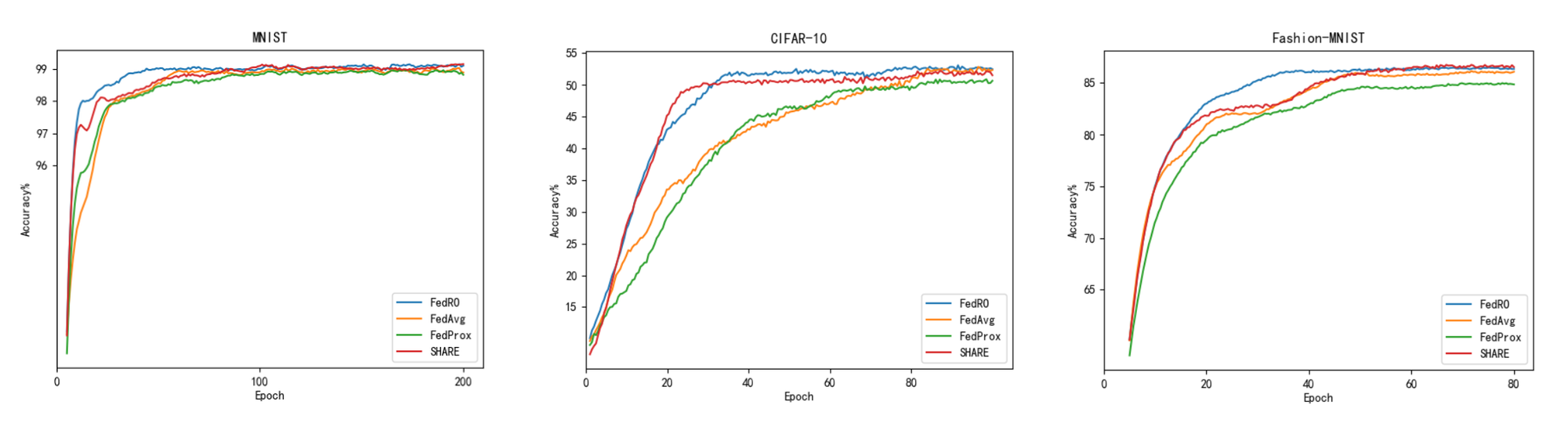}
    \caption{Experimental results on the MNIST, CIFAR-10, and Fashion-MNIST datasets.}
    \Description{Three line charts show the accuracy curves of FedAvg, FedProx, SHARE, and FedRO on MNIST, CIFAR-10, and Fashion-MNIST.}
    \label{fig:results}
\end{figure*}

Figure~\ref{fig:results} shows the experimental results of different algorithms using three datasets when $\sigma=0.8$. Table~\ref{tab:highest-accuracy} shows the maximum accuracy that three algorithms can achieve on each dataset. In Table~\ref{tab:rounds}, each entry shows the number of communication rounds required to achieve a test set accuracy of 99\% on MNIST, 55\% on CIFAR-10, and 85\% on Fashion-MNIST for CNN. In this article, we conducted experiments at $\sigma=0.8$ and 1.0, and counted the number of communication rounds required for different algorithms to achieve the standard accuracy. The shorter the number of rounds, the more efficient the algorithm is, the lower the workload, and the higher the performance.

\begin{table*}[t]
\centering
\caption{The highest accuracy that can be achieved.}
\label{tab:highest-accuracy}
\begin{tabularx}{\textwidth}{L Y Y Y}
\toprule
Algorithm & MNIST & CIFAR-10 & Fashion-MNIST \\
\midrule
FedAvg & 0.991 & 0.557 & 0.872 \\
FedProx & 0.993 & 0.552 & 0.863 \\
SHARE & \textbf{0.995} & 0.564 & \textbf{0.879} \\
FedRO(Ours) & 0.993 & \textbf{0.576} & 0.874 \\
\bottomrule
\end{tabularx}
\end{table*}

\begin{table*}[t]
\centering
\caption{The number of communication rounds to reach a target accuracy.}
\label{tab:rounds}
\begin{tabularx}{\textwidth}{L Y Y Y Y}
\toprule
Algorithm & $\sigma$ & MNIST & CIFAR-10 & Fashion-MNIST \\
\midrule
\multirow{2}{*}{FedAvg} & 1.0 & 1517 & 1714 & 1811 \\
& 0.8 & 221 & 87 & 52 \\
\midrule
\multirow{2}{*}{FedProx} & 1.0 & 1421 & 1658 & 1720 \\
& 0.8 & 210 & 96 & 70 \\
\midrule
\multirow{2}{*}{SHARE} & 1.0 & 1201 & 1320 & \textbf{1526} \\
& 0.8 & \textbf{142} & 76 & 48 \\
\midrule
\multirow{2}{*}{FedRO(Ours)} & 1.0 & \textbf{1105} & \textbf{1240} & 1534 \\
& 0.8 & 152 & \textbf{70} & \textbf{46} \\
\bottomrule
\end{tabularx}
\end{table*}

From the experimental results, it can be seen that FedRO has good performance on all three datasets. The experimental results of this article show that FedRO can reduce the number of communications by up to 27\% on the MNIST dataset, up to 38\% on the CIFAR-10 dataset, and up to 17\% on the Fashion-MNIST dataset.

\subsection{Experimental Results and Analysis of Clustering Algorithms}

Table~\ref{tab:clustering} shows the clustering performance of four clustering algorithms on four datasets, including the values of ARI and NMI. The closer the value is to 1, the better the clustering effect of the algorithm.

\begin{table*}[t]
\centering
\caption{Experimental results on ARI and NMI.}
\label{tab:clustering}
\begin{tabularx}{\textwidth}{L Y Y Y Y Y}
\toprule
Algorithm & Parameter & glass & wine & yeast & iris \\
\midrule
\multirow{2}{*}{K-means} & ARI & 0.802 & 0.304 & 0.386 & 0.547 \\
& NMI & 0.745 & 0.241 & \textbf{0.402} & 0.681 \\
\midrule
\multirow{2}{*}{OPTICS} & ARI & 0.293 & 0.311 & 0.258 & 0.561 \\
& NMI & 0.391 & 0.294 & 0.243 & \textbf{0.758} \\
\midrule
\multirow{2}{*}{DBSCAN} & ARI & \textbf{0.838} & 0.293 & 0.375 & 0.535 \\
& NMI & 0.732 & 0.335 & 0.336 & 0.684 \\
\midrule
\multirow{2}{*}{FedRO(Ours)} & ARI & 0.792 & \textbf{0.334} & \textbf{0.415} & \textbf{0.612} \\
& NMI & \textbf{0.852} & \textbf{0.372} & 0.305 & 0.624 \\
\bottomrule
\end{tabularx}
\end{table*}

From the experimental results, it can be seen that the reinforcement learning-based adaptive OPTICS clustering algorithm proposed in this article has good clustering performance on all four datasets. The algorithm dynamically adjusts the clustering coefficient by using the NMI value as the output of the reward function, which leads to good NMI values in the clustering task. In contrast, the non-adaptive OPTICS clustering algorithm has poor clustering performance because it cannot effectively handle datasets with no clear boundaries between clusters, and it cannot effectively cluster high-dimensional datasets. However, the proposed clustering algorithm in this article can achieve good results even on high-dimensional datasets by adaptively adjusting parameters.

\section{Conclusion}

This article proposes a reinforcement learning-based federated learning method based on adaptive OPTICS clustering, which can solve the heterogeneity problem of data distribution in federated learning tasks. Compared with traditional federated learning algorithms, the proposed algorithm can not only achieve efficient data recording, learning, and updating but also share and aggregate data while protecting user privacy. In the case of data heterogeneity, it reduces the impact caused by different data distributions, improves accuracy, and significantly improves performance.

When clustering data, there needs to be a balance between the number of clusters. The more clusters there are, the higher the accuracy of the models within each cluster, but the worse the global model's generalization will be. However, if there are too few clusters, the accuracy of the models within each cluster will decrease. By using adaptive parameters, we can determine a range that both clusters the data in a better range and reduces the number of iterations required for adaptive parameter adjustment. In future research, we will further optimize performance through further study and experimentation.

\begin{acks}
This work was supported by the National Natural Science Foundation of China (62192784, U22B2038, 62172056).
\end{acks}

\bibliographystyle{tianyu_custom_unsrt}
\bibliography{references}

\end{document}